\documentclass{article}

\usepackage{microtype}
\usepackage{graphicx}
\usepackage{subfigure}
\usepackage{booktabs} % for professional tables
\usepackage{enumitem}
\usepackage{multicol}
\usepackage{wrapfig}

\usepackage{algorithm}
\usepackage{algorithmic}
\usepackage{bbm}
\usepackage{acronym}
\usepackage{amssymb}
\usepackage{amsmath}
\usepackage[title]{appendix}
\usepackage{multirow}
\usepackage{dblfloatfix}
\usepackage{tikz}
\usepackage{pgfplots}
\usepackage[font={small}]{caption}
\usepgfplotslibrary{patchplots}
\usetikzlibrary{patterns, positioning, arrows}
\usetikzlibrary{shapes, arrows.meta}

\acrodef{RL}{Reinforcement Learning}
\acrodef{DR}{Domain Randomization}
\acrodef{VDR}{Vanilla Domain Randomization}
\acrodef{DRL}{Deep Reinforcement Learning}
\acrodef{ADR}{Active Domain Randomization}
\acrodef{MDP}{Markov Decision Process}
\acrodef{SVGD}{Stein Variational Gradient Descent}
\acrodef{SVPG}{Stein Variational Policy Gradient}
\acrodef{MES}{main engine strength}
\acrodef{DoF}{degrees of freedom}
\acrodef{RBF}{Radial Basis Function}
\acrodef{A2C}{Advantage Actor-Critic}
\acrodef{UDR}{Uniform Domain Randomization}
\acrodef{BO}{Bayesian Optimization}
\acrodef{RARL}{Robust Adversarial Reinforcement Learning}

\newcommand\blfootnote[1]{%
  \begingroup
  \renewcommand\thefootnote{}\footnote{#1}%
  \addtocounter{footnote}{-1}%
  \endgroup
}

\usepackage{hyperref}
\usepackage[final]{corl_2019} 
\allowdisplaybreaks[1]

\usepackage{xcolor}
\title{Active Domain Randomization}

\author{
  Bhairav Mehta\\
  Mila, Universit\'{e} de Montr\'{e}al\\
  \And 
  Manfred Diaz\\
  Mila, Universit\'{e} de Montr\'{e}al\\
  \And
  Florian Golemo\\
  Mila, Universit\'{e} de Montr\'{e}al, ElementAI\\
  \AND
  Christopher J. Pal\\
  Mila, Polytechnique Montr\'{e}al, ElementAI, CIFAR\\
  \And
  Liam Paull\\
  Mila, Universit\'{e} de Montr\'{e}al, CIFAR
}

\begin{document}
\maketitle

\begin{abstract}

Domain randomization is a popular technique for improving domain transfer, often used in a zero-shot setting when the target domain is unknown or cannot easily be used for training. In this work, we empirically examine the effects of domain randomization on agent generalization. Our experiments show that domain randomization may lead to suboptimal, high-variance policies, which we attribute to the uniform sampling of environment parameters. We propose Active Domain Randomization, a novel algorithm that learns a parameter sampling strategy. Our method looks for the most informative environment variations within the given randomization ranges by leveraging the discrepancies of policy rollouts in randomized and reference environment instances. We find that training more frequently on these instances leads to better overall agent generalization. 
Our experiments across various physics-based simulated and real-robot tasks show that this enhancement leads to more robust, consistent policies.

\end{abstract}

\keywords{sim2real, domain randomization, reinforcement learning}

\blfootnote{Correspondence to \texttt{mehtabha@mila.quebec}}

\section{Introduction}
Recent trends in \ac{DRL} exhibit a growing interest in zero-shot domain transfer, i.e. when a policy is learned in a source domain and is then tested \textit{without finetuning} in a previously unseen target domain. 
Zero-shot transfer is particularly useful when the task in the target domain is inaccessible, complex, or expensive, such as gathering rollouts from a real-world robot. 
An ideal agent would learn to \textit{generalize} across domains; it would accomplish the task without exploiting irrelevant features or deficiencies in the source domain (i.e., approximate physics in simulators), which may vary dramatically after transfer. 

One promising approach for zero-shot transfer has been \ac{DR} \citep{tobin2017domain}. 
In \ac{DR}, we uniformly randomize environment parameters (i.e. friction, motor torque) in predefined ranges after every training episode. 
By randomizing everything that might vary in the target environment, the hope is that the agent will view the target domain as just another variation. 
However, recent works suggest that the sample complexity grows exponentially with the number of randomization parameters, even when dealing only with transfer between simulations (i.e. in \citet{openai2018learning} Figure 8). %Note: refer directly to Figure 8
In addition, when using \ac{DR} \textit{unsuccessfully}, policy transfer fails, but with no clear way to understand the underlying cause. 
After a failed transfer, randomization ranges are tweaked heuristically via trial-and-error. 
Repeating this process iteratively leads to arbitrary ranges that do (or do not) lead to policy convergence without any insight into how those settings may affect the learned behavior.

\begin{figure}[tb]
\begin{minipage}[c]{0.6\columnwidth}
    \centerline{\includegraphics[width=0.95\columnwidth]{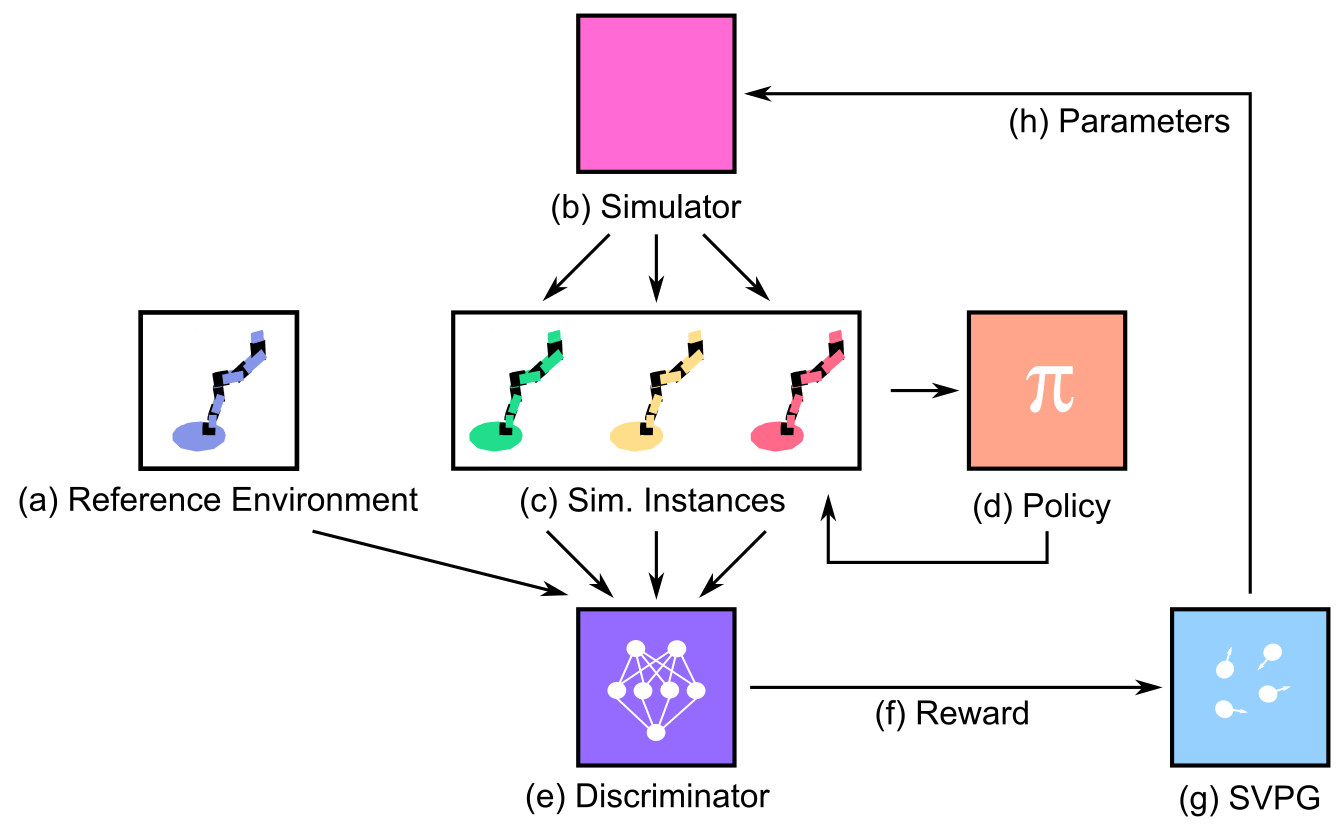}}
\end{minipage}\hfill
\begin{minipage}[c]{0.4\columnwidth}
    \caption{ADR proposes \textbf{randomized environments} (c) or simulation instances from a  \textbf{simulator} (b) and rolls out  an \textbf{agent policy} (d) in those instances. The \textbf{discriminator} (e) learns a \textbf{reward} (f) as a proxy for environment difficulty by distinguishing between rollouts in the \textbf{reference environment} (a) and randomized instances, which is used to train \textbf{SVPG particles} (g). The particles propose a diverse set of environments, trying to find the environment \textbf{parameters} (h) that are currently causing the agent the most difficulty.
    }
    \label{fig:overview}
\end{minipage}

% (a) \textbf{Reference Environment}, we assume that we have access to a (b) \textbf{Simulator} with parameterizable dynamics, along with a set of default settings for environment dynamics; (c) \textbf{Randomized Instances}, the environment instances that are the result of specific environment dynamics. 
% In practice, we usually have 10-15 instances in parallel; 
% (d) \textbf{Policy}, which can be any RL algorithm that interacts with the current set of simulator instances to update the policy; 
% (e) \textbf{Discriminator}, which receives trajectories from the policy in the reference environment and in the randomized instances and learns to distinguish the two as a proxy for environment difficulty; (f) \textbf{Reward}, the output of discriminator is used train the SVPG particles; (g) \textbf{Stein Variational Policy Gradient (SVPG) Sampler}, a policy which maintains a set of particles, each of which proposes a set of environment dynamics. Induces training environment variety with repulsive kernel term; (h) \textbf{Parameters}, each particle's proposed instances are sent to the simulator to create a new randomized instance.}
%\label{fig:overview}
% \end{center}
\end{figure}

In this work, we demonstrate that the strategy of \textit{uniformly} sampling environment parameters from predefined ranges is suboptimal and propose an alternative sampling method, \textbf{Active Domain Randomization}. 
\ac{ADR}, shown graphically in Figure~\ref{fig:overview}, formulates \ac{DR} as a search for randomized environments that maximize utility for the agent policy. 
Concretely, we aim to find environments that \textit{currently} cause difficulties for the agent policy, dedicating more training time to these troublesome parameter settings. 
We cast this active search as a \ac{RL} problem where the \ac{ADR} sampling policy is parameterized with \ac{SVPG} \cite{svpg}. 
\ac{ADR} focuses on problematic regions of the randomization space by learning a discriminative reward computed from discrepancies in policy rollouts generated in randomized and reference environments.

We first showcase \ac{ADR} on a simple environment where the benefits of training on more challenging variations are apparent and interpretable (Figure \ref{fig:ll-generalization}). In this case, we demonstrate that \ac{ADR} learns to preferentially select parameters from  these more challenging parameter regions while still adapting to the policy's current deficiencies.
We then apply \ac{ADR} to more complex environments and real robot settings (Figure \ref{fig:realrobots}) and show that even with high-dimensional search spaces and unmodeled dynamics, policies trained with \ac{ADR} exhibit superior generalization and lower overall variance than their \ac{UDR} counterparts.

% Our proposed \ac{ADR} method learns an adaptive randomization strategy that finds problematic environments within the given randomization ranges. Across a wide variety of simulated and real-world tasks, we find that training preferentially on these environments leads to better generalization.
    % \item \ac{ADR} can provide insight into which dimensions and parameter ranges are most influential \textit{before transfer}, which can aid the tuning of randomization ranges before expensive experiments are undertaken. 
% \end{enumerate}

\section{Preliminaries}

% In this section, we briefly cover the basics of \ac{RL} (used to train both the agent policy and the \ac{ADR} policy), Stein Variational Policy Gradient (parameterizes the \ac{ADR} policy), and domain randomization.

\subsection{Reinforcement Learning}
We consider a \ac{RL} framework \cite{sutton2018reinforcement} where some task $T$ is defined by a \ac{MDP} consisting of a state space $S$, action space $A$, state transition function $P: S \times A \mapsto S$, reward function $R: S \times A \mapsto \mathbb{R}$, and discount factor $\gamma \in (0,1)$. 
The goal for an agent trying to solve $T$ is to learn a policy $\pi$ with parameters $\theta$ that maximizes the expected total discounted reward.
We define a rollout $\tau = (s_0, a_0 ..., s_T, a_T)$ to be the sequence of states $s_t$ and actions $a_t \sim \pi(a_t | s_t)$ executed by a policy $\pi$ in the environment. 

\subsection{Stein Variational Policy Gradient}
Recently, \citet{svpg} proposed \ac{SVPG}, which learns an ensemble of policies $\mu_\phi$ in a maximum-entropy \ac{RL} framework \cite{ziebart2010modeling}.
\begin{equation}
    \max_\mu \mathbb{E}_\mu[J(\mu)] + \alpha \mathcal{H}(\mu)
\end{equation}

with entropy $\mathcal{H}$ being controlled by temperature parameter $\alpha$. \ac{SVPG} uses Stein Variational Gradient Descent \cite{svgd} to iteratively update an ensemble of $N$ policies or \textit{particles} $\mu_\phi = \{\mu_{\phi_i}\}_{i=1}^N$ using:

\begin{equation}
   \mu_{\phi_i} \leftarrow \mu_{\phi_i} + 
    \frac{\epsilon}{N}\sum_{j=1}^N[\nabla_{\mu_{\phi_j}}J(\mu_{\phi_j})k(\mu_{\phi_j}, \mu_{\phi_i}) +  \alpha\nabla_{\mu_{\phi_j}}k(\mu_{\phi_j}, \mu_{\phi_i})]
    \label{eq:svpg}
\end{equation}

with step size $\epsilon$ and positive definite kernel $k$. 
This update rule balances exploitation (first term moves particles towards high-reward regions) and exploration (second term repulses similar policies). 

\subsection{Domain Randomization}
Domain randomization (DR) is a technique to increase the generalization capability of policies trained in simulation. 
\ac{DR} requires a prescribed set of $N_{rand}$ simulation parameters to randomize, as well as corresponding ranges to sample them from. 
A set of parameters is sampled from \textit{randomization space} $\Xi \subset \mathbb{R}^{N_{rand}}$, where each randomization parameter $\xi^{(i)}$ is bounded on a closed interval $\{ \big[ \xi^{(i)}_{low}, \xi^{(i)}_{high} \big] \}_{i=1}^{N_{rand}}$. 

When a configuration $\xi \in \Xi$ is passed to a non-differentiable simulator $S$, it generates an environment $E$. At the start of each episode, the parameters are uniformly sampled from the ranges, and the environment generated from those values is used to train the agent policy $\pi$. 

\ac{DR} may perturb any to all elements of the task $T$'s underlying \ac{MDP}\footnotemark, with the exception of keeping $R$ and $\gamma$ constant. \ac{DR} therefore generates a set of \ac{MDP}s that are superficially similar, but can vary greatly in difficulty depending on the character of the randomization.
Upon transfer to the target domain, the expectation is that the agent policy has learned to generalize across \ac{MDP}s, and sees the final domain as just another variation of parameters. 

The most common instantiation of \ac{DR}, \ac{UDR} is summarized in Algorithm \ref{alg:dr} in Appendix \ref{app:udr}.
\ac{UDR} generates randomized environment instances $E_i$ by uniformly sampling $\Xi$.
The agent policy $\pi$ is then trained on rollouts ${\tau_i}$ produced in randomized environments ${E_i}$.

\footnotetext{The effects of \ac{DR} on action space $A$ are usually implicit or are carried out on the simulation side.}

\section{Method}
\label{sec:method}

To start, we would like to answer the following question:

\begin{center}
    \textit{Are all MDPs generated by uniform randomization equally useful for training?}
\end{center}

We consider the \texttt{LunarLander-v2} environment, where the agent's task is to ground a lander in a designated zone and reward is based on the quality of landing (fuel used, impact velocity, etc).
\texttt{LunarLander-v2} has one main axis of randomization that we vary: the \ac{MES}.

We aim to determine if certain environment instances (different values of the \ac{MES}) are more \textit{informative} - more efficient than others - in terms of aiding generalization. 
We set the total range of variation for the \ac{MES} to be $[ 8, 20]$\footnotemark\footnotetext{Default MES is 13; MES $\leq$ 7.5 is unsolvable when all other parameters remain constant.} and find through empirical tests that lower engine strengths generate harder \ac{MDP}s to solve. 
Under this assumption, we show the effects of \textit{focused \ac{DR}} by editing the range that the \ac{MES} parameter is uniformly sampled from.
\begin{wrapfigure}{l}{0.5\columnwidth}
    \centering
    \includegraphics[width=.45\columnwidth]{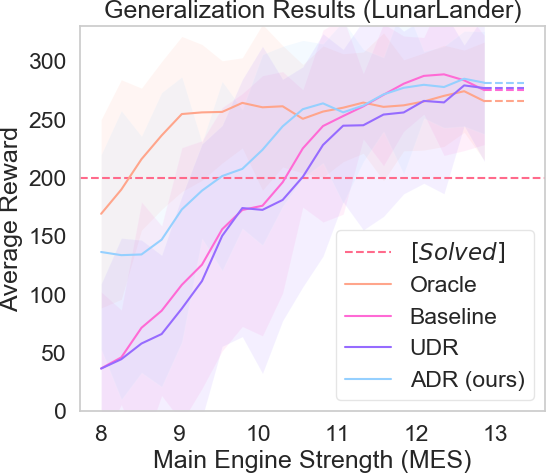} 
    % \caption{This plot shows the ability of different training methods to transfer and generalize over environment settings. We compared the following approaches: Baseline, i.e. default environment dynamics; Uniformly sampled Domain Randomization (UDR); Active Domain Randomization (ADR, our approach in which we actively search for difficult environment dynamics); and Oracle, i.e. a handpicked randomization range. For the evaluation, we took policies that were trained with these methods and evaluated them on a wide range of environment parameters (i.e. main engine strength in this "LunarLander" environment which affects the responsiveness and landing speed of the simulated landing robot). }
    \caption{Agent generalization, expressed as performance across different engine strength settings in \textbf{LunarLander}. 
    We compare the following approaches: Baseline (default environment dynamics); Uniform Domain Randomization (UDR); Active Domain Randomization (ADR, our approach); and Oracle (a handpicked randomization range of MES ~ [8, 11]). 
    % For evaluation, we take each sampling strategy's final policies and evaluate them across the full range of environment parameters (i.e. vary main engine strength, which affects the responsiveness and landing speed of the simulated lander). 
    ADR achieves for near-expert levels of generalization, while both Baseline and UDR fail to solve lower MES tasks.
    }
    \label{fig:ll-generalization}
    \vspace{-30pt}
\end{wrapfigure}

We train multiple agents on different randomization ranges for \ac{MES}, which define what types of environments the agent is exposed to during training. Figure \ref{fig:ll-generalization} shows the final generalization performance of each agent by sampling randomly from the entire randomization range of [8, 20] and rolling out the policy in the generated environments. We see that, in this case, focusing on harder \ac{MDP}s improves generalization as compared to uniformly sampling the whole space, even when the evaluation environment is outside of the training distribution.

\subsection{Problem Formulation}
The experiment in the previous section shows that preferential training on more informative environments provides tangible benefits in terms of agent generalization. However, in general, finding these informative environments is difficult because: \textbf{(1)} It is rare that such intuitively \textit{hard} \ac{MDP} instances or parameter ranges are known beforehand and \textbf{(2)} \ac{DR} is used mostly when the space of randomized parameters is high-dimensional or noninterpretable. 
As a result, we propose an algorithm for finding environment instances that maximize \textit{utility}, or provide the most improvement (in terms of generalization) to our agent policy when used for training.

% To formalize the search problem described, we introduce the following \textit{generalization objective}:

% \begin{equation}
% \label{eq:generalizationobj}
%     \arg \max_{\{ \xi_i \}_{i=1}^N \in \Xi_{train}} |S_{test}|
% \end{equation}

% where $S_{test}$ is the set of solved, potentially unseen, task variations from $ \Xi_{test}$. To compute this quantity, we roll out policy $\pi$, trained on $N$ environments {$\{ \xi_i \}_{i=1}^N \in \Xi_{train}$}, and evaluate it on environments sampled from $\Xi_{test}$. We would like to see if the policy exceeds a pre-determined \textit{Solved} score $r_{solved}$, specified by the task definition.

% \begin{equation}
%     S_{test} = \{\xi | \mathbbm{1}[\mathbb{E}_\pi[J(\theta)] > r_{solved}, \enskip \forall \xi \in \Xi_{test}] \}
% \end{equation}

% In realistic scenarios such as sim2real transfer, it is likely that $\Xi_{train} \cap \Xi_{test} = \emptyset$. Given access only to the former, we'd like to choose environments $\{ \xi_i \}_{i=1}^N \in \Xi_{train}$ that when used to train a policy, maximize the \textit{generalization performance} when evaluated on environments $\xi \in \Xi_{test}$.

\subsection{Active Domain Randomization}
\label{sec:adr}

Drawing analogies with \ac{BO} literature, one can consider the randomization space as a search space.
Traditionally, in \ac{BO}, the search for where to evaluate an objective is informed by acquisition functions, which trade off exploitation of the objective with exploration in the uncertain regions of the space \cite{brochubayesopt}. 
However, unlike the stationary objectives seen in \ac{BO}, training the agent policy renders our optimization non-stationary: the environment with highest utility at time $t$ is likely not the same as the maximum utility environment at time $t+1$.
This requires us to redefine the notion of an acquisition function while simultaneously dealing with \ac{BO}'s deficiencies with higher-dimensional inputs \cite{wangrembo2013}. 
% With this dynamic objective, we need to actively search the space for the most fruitful training environments given the current state of the agent policy. 

% \begin{enumerate}[noitemsep, nosep]
%     \item It is rare that such intuitively \textit{hard} \ac{MDP} instances or parameter ranges are known beforehand.
%     \item \ac{DR} is used mostly when the space of randomized parameters is high-dimensional or noninterpretable.
% \end{enumerate}

% An ideal randomization scheme would find the most informative environment instances in the randomization space, rather than uniformly sampling from the entire space. While seemingly just an instantiation of the traditional \ac{BO} problem, the nonstationarity of the objective (the environment utility) 

\noindent
\begin{wrapfigure}{L}{0.6\textwidth}
\vspace{-20pt}
\begin{minipage}[c]{0.6\columnwidth}
    \begin{algorithm}[H]
       \caption{Active Domain Randomization}
       \label{alg:adr}
    \begin{algorithmic}[1]
       \STATE {\textbf{Input}: $\Xi$: Randomization space, $S$: Simulator, $\xi_{ref}$: reference parameters} 
       \STATE \textbf{Initialize} $\pi_\theta$: agent policy, $\mu_{\phi}$: SVPG particles, $D_{\psi}$: discriminator,  $E_{ref} \leftarrow S(\xi_{ref})$: reference environment
       \WHILE{ \textbf{not} $max\_timesteps$}
       \FOR{ \textbf{each} particle $\mu_\phi$}
       \STATE \textbf{rollout} $\xi_i \sim \mu_\phi(\cdot)$
       \ENDFOR
       \FOR{\textbf{each} $\xi_i$}
       \STATE { \textit{// Generate, rollout in randomized env.}} \label{l:start-rollout}
       \STATE $E_i \leftarrow S(\xi_i)$ 
       \STATE \textbf{rollout} $\tau_i \sim \pi_\theta(\cdot; E_{i})$,  $\tau_{ref} \sim \pi_\theta(\cdot; E_{ref})$ 
       \STATE{$\mathcal{T}_{rand} \gets \mathcal{T}_{rand} \cup \tau_i$; 
       $\mathcal{T}_{ref} \gets \mathcal{T}_{ref} \cup \tau_{ref}$}
       \ENDFOR \label{l:end-rollout}
       \STATE \textit{// Calculate reward for each proposed environment}
       \FOR{\textbf{each} $\tau_i \in \mathcal{T}_{rand}$}
       \STATE{Calculate $r_i$ for $\xi_i$ / $E_i$ (Eq. (\ref{eq:discrimrew}))}
       \ENDFOR
       \STATE \textit{// Gradient Updates}
       \STATE \textbf{with} $\mathcal{T}_{rand}$ \textbf{update}:
       \STATE $\quad \theta \leftarrow \theta + \nu\nabla_{\theta}J(\pi_\theta)$
       \STATE Update particles using Eq. (\ref{eq:svpg})
       \STATE Update $D_\psi$ with $\tau_i$ and $\tau_{ref}$ using SGD.
       \ENDWHILE
       
    \end{algorithmic}
    \end{algorithm}
    \vspace{-20pt}
\end{minipage}%NO Space!!!
\end{wrapfigure}
% \begin{minipage}[c]{0.35\columnwidth}% 
% \centering
% \includegraphics[width=.6\columnwidth]{plots2/ergo-reacher-sim.png}
% %   \quad 
% \includegraphics[width=.6\columnwidth]{plots2/ergo-reacher-real.png}
% \captionof{plots2/ergo-reacher-real.png}{ADR shows benefits over UDR on a wide range of tasks, including in a sim2real reaching task. The 4 DoF simulated robot must learn an efficient policy to reach a virtual point (shown in pink), and the final policies are evaluated on the real robot. We show that ADR policies transfer more robustly during zero-shot transfer to the more difficult real-world robot environment.}
% % \caption{We've evaluated our approach on this sim2real task (among others), in which the 4 DoF simulated robot has to learn a policy  to reach a virtual point (here in pink) as soon as possible and the policies are evaluated on the real robot. This task is based on the famous OpenAI Gym environment "Reacher", but with the real-world robotic implementation and higher difficulty. } 
%   \label{fig:ergosimreal}
% \end{minipage}

To this end, we propose \ac{ADR}, summarized in Algorithm \ref{alg:adr} and Figure \ref{fig:overview}\footnotemark.
\footnotetext{We provide a detailed walkthrough of the algorithm in Appendix \ref{sec:walkthrough}.} \ac{ADR} provides a framework for manipulating a more general analog of an \textit{acquisition function}, selecting the most informative \ac{MDP}s for the agent within the randomization space. 
By formulating the search as an \ac{RL} problem, \ac{ADR} learns a policy $\mu_{\phi}$ where states are proposed randomization configurations $\xi \in \Xi$ and actions are continuous changes to those parameters.

We learn a discriminator-based reward for $\mu_{\phi}$, similar the reward seen in \citet{eysenbach2018diversity}:
\begin{equation}
    r_{D} = \log D_{\psi}(y|\tau_{i} \sim \pi(\cdot; E_{i}))
\label{eq:discrimrew}
\end{equation}
where $y$ is a boolean variable denoting the discriminator's prediction of which type of environment (a randomized environment $E_{i}$ or reference environment $E_{ref}$) the trajectory $\tau_{i}$ was generated from. 
We assume that the $E_{ref} = S(\xi_{ref})$ is provided with the original task definition.

Intuitively, we reward the policy $\mu_\phi$ for finding regions of the randomization space that produce environment instances where the \textit{same} agent policy $\pi$ acts differently than in the reference environment.
The agent policy $\pi$ sees and trains \textit{only} on the randomized environments (as it would in traditional \ac{DR}), using the environment's task-specific reward for updates.
As the agent improves on the proposed, problematic environments, it becomes more difficult to differentiate whether any given state transition was generated from the reference or randomized environment. 
Thus, \ac{ADR} can find what parts of the randomization space the agent is currently performing poorly on, and can \textit{actively} update its sampling strategy throughout the training process.

\section{Results}
\subsection{Experiment Details}\label{implementation}
\begin{figure}[h]
  \centering
  \subfigure[]{% 
    \includegraphics[width=.21\columnwidth]{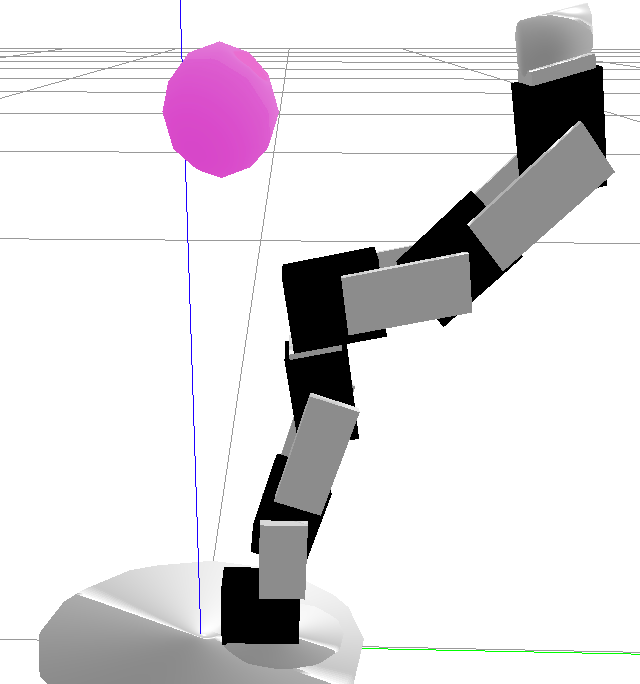}
  } 
  ~
  \subfigure[]{% 
    \includegraphics[width=.21\columnwidth]{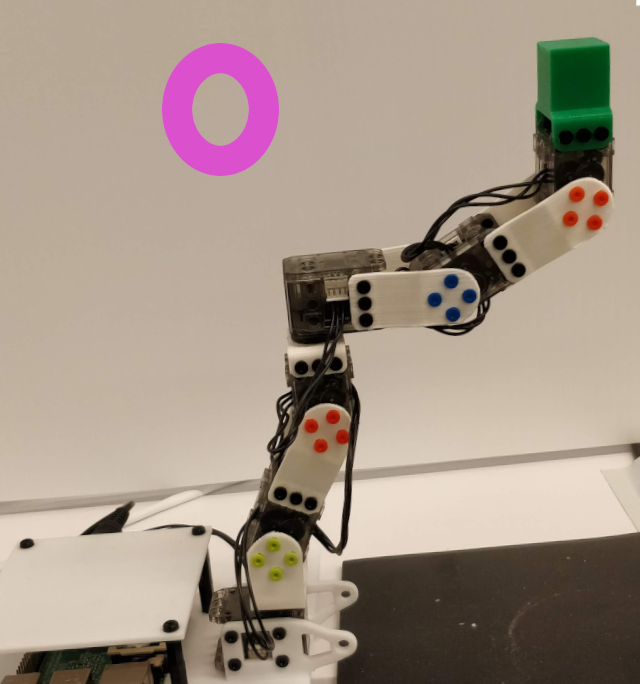} 
  } 
  ~ 
  \subfigure[]{% 
    \includegraphics[width=.225\columnwidth]{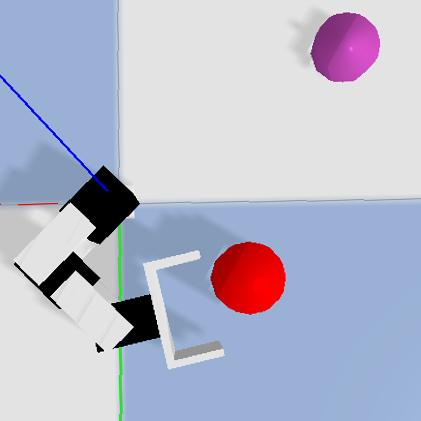}
  } 
  ~
  \subfigure[]{% 
    \includegraphics[width=.225\columnwidth]{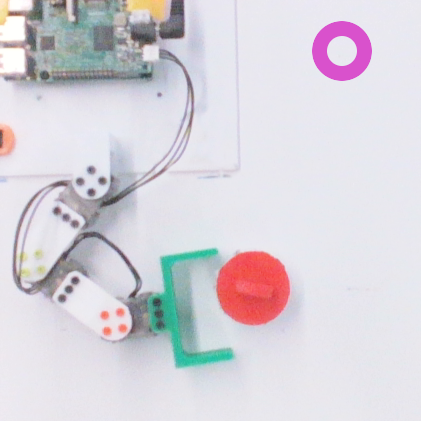}
  } 
  \caption{Along with simulated environments, we display ADR on zero-shot transfer tasks onto real robots.} 
%   \label{lunarours}
% \end{figure}
% \begin{figure}[t]
%   \caption{ \textbf{(b)} Change in dynamics sampling during training. As training proceeds, ADR begins preferentially sampling the more challenging environmental instances.} 
  \label{fig:realrobots}
\end{figure}

To test ADR, we experiment on OpenAI Gym environments \cite{brockman2016gym} across various tasks, both simulated and real:
\textbf{(a)}  \texttt{LunarLander-v2}, a 2 \ac{DoF} environment where the agent has to softly land a spacecraft, implemented in Box2D (detailed in Section \ref{sec:adr}), 
\textbf{(b)} \texttt{Pusher-3DOF-v0}, a 3 \ac{DoF} arm from \citet{haarnoja2018composable} that has to push a puck to a target, implemented in Mujoco \cite{mujoco2012},
and \textbf{(c)} \texttt{ErgoReacher-v0}, a 4 \ac{DoF} arm from \citet{golemo2018sim} which has to touch a goal with its end effector, implemented in the Bullet Physics Engine \cite{bulletphys2015}. For sim2real experiments, we recreate this environment setup on a real Poppy Ergo Jr. robot \cite{lapeyre:tel-01104641} shown in Figure\ref{fig:realrobots} (a) and (b),
and also create \textbf{(d)} \texttt{ErgoPusher-v0} an environment similar to \texttt{Pusher-3DOF-v0} with a real robot analog seen in Figure \ref{fig:realrobots} (c) and (d). 
We provide a detailed account of the randomized parameters in each environment in Table \ref{table:envtable} in Appendix \ref{app:envtable}.

All simulated experiments are run with five seeds each with five random resets, \textbf{totaling 25 independent trials per evaluation point}. All experimental results are plotted mean-averaged with one standard deviation shown. 
Detailed experiment information can be found in Appendix \ref{app:expdetails}. 

\subsection{Toy Experiments}

To investigate whether \ac{ADR}'s learned sampling strategy provides a tangible benefit for agent generalization, we start by comparing it against traditional \ac{DR} (labeled as \ac{UDR}) on \texttt{LunarLander-v2} and vary only the main engine strength (MES). In Figure \ref{fig:ll-generalization}, we see that \ac{ADR} approaches expert-levels of generalization whereas \ac{UDR} fails to generalize on lower \ac{MES} ranges.

\begin{figure}[h]
  \centering
  \subfigure[]{% 
    \includegraphics[width=.3\columnwidth]{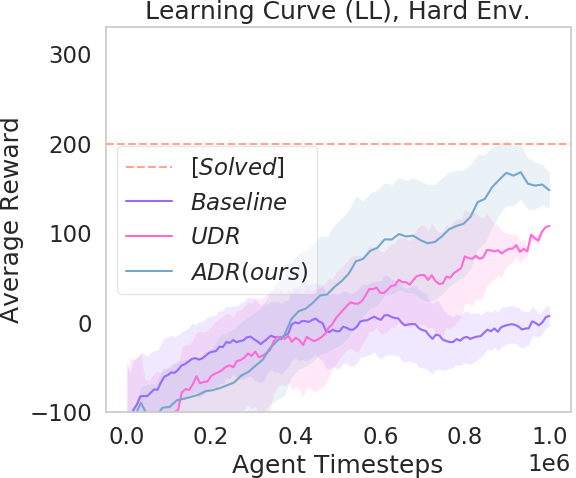}\label{lunarhard} 
  } 
  ~
  \subfigure[]{% 
    \includegraphics[width=.3\columnwidth]{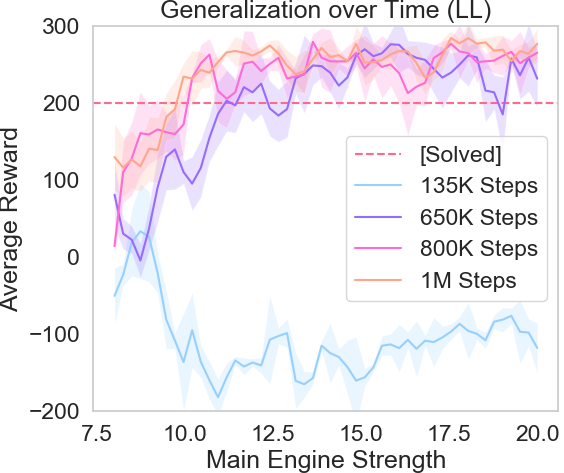} \label{evolutiongen} 
  } 
  ~ 
  \subfigure[]{% 
    \includegraphics[width=.3\columnwidth]{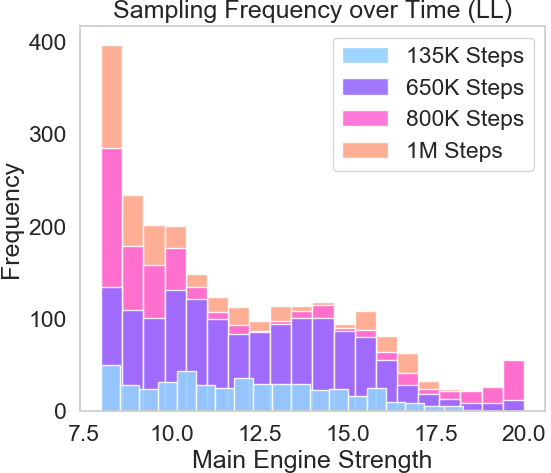}\label{evolutionsamping} 
  } 
  \caption{Learning curves over time in \textbf{LunarLander}. Higher is better. \textbf{(a)} Performance on particularly difficult settings - our approach outperforms both the policy trained on a single simulator instance ("baseline") and the UDR approach. \textbf{(b)} Agent generalization in \textbf{LunarLander} over time during training when using ADR. \textbf{(c)} Adaptive sampling visualized. ADR, seen in \textbf{(b)} and \textbf{(c)}, adapts to where the agent is struggling the most, improving generalization performance by end of training.} 
%   \label{lunarours}
% \end{figure}
% \begin{figure}[t]
%   \caption{ \textbf{(b)} Change in dynamics sampling during training. As training proceeds, ADR begins preferentially sampling the more challenging environmental instances.} 
  \label{evolution}
\end{figure}

% From Figure \ref{lunarref}, we see that \ac{ADR} solves the reference environment ($\xi_{MES}=13$) more consistently than \ac{UDR}, never dipping below the \textit{Solved} line once that level of performance is reached.
% Figure  shows that 
We compare the learning progress for the different methods on the \textit{hard} environment instances ($\xi_{MES} \sim U[8, 11]$) in Figure \ref{lunarhard}. \ac{ADR} significantly outperforms both the baseline (trained only on \ac{MES} of 13) and the \ac{UDR} agent (trained seeing environments with  $\xi_{MES} \sim U[8, 20]$) in terms of performance.

Figures \ref{evolutiongen} and \ref{evolutionsamping} showcase the adaptability of \ac{ADR} by showing generalization and sampling distributions at various stages of training. 
\ac{ADR} samples approximately uniformly for the first 650K steps, but then finds a deficiency in the policy on higher ranges of the \ac{MES}. 
As those areas become more frequently sampled between 650K-800K steps, the agent learns to solve all of the higher-MES environments, as shown by the generalization curve for 800K steps. As a result, the discriminator is no longer able to differentiate reference and randomized trajectories from the higher MES regions, and starts to reward environment instances generated in the lower end of the MES range, which improves generalization towards the completion of training.

\subsection{Randomization in High Dimensions}
\begin{figure}[t]
\centering
  \subfigure[]{% 
    \includegraphics[width=0.475\columnwidth]{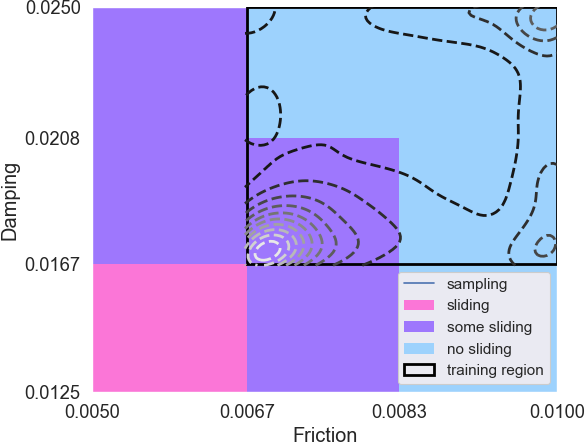} \label{pushersampling} 
  } 
  ~ 
  \subfigure[]{% 
    \includegraphics[width=0.475\columnwidth]{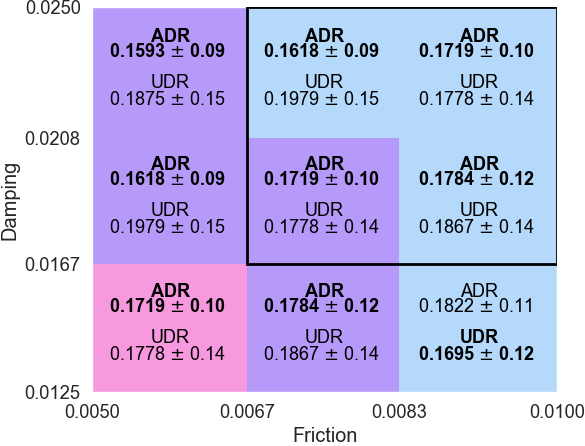} \label{pusheradvantage} 
  } 
  \caption{In \textbf{Pusher-3Dof}, the environment dynamics are characterized by friction and damping of the sliding puck, where sliding correlates with the difficulty of the task (as highlighted by cyan, purple, and pink - from easy to hard). \textbf{(a)} During training, the algorithm only had access to a limited, easier range of dynamics (black outlined box in the upper right). 
%   We observed that our approach will converge to the hardest settings within this limited range.
  \textbf{(b)} Performance measured by distance to target, lower is better. }
%   The results show the higher performance and lower variance of our approach, safe for one exception.}
%   Agent generalization (b) and environment sampling frequency (a) throughout training on \texttt{ErgoPusher-v0}.
  
  \label{pushergrid}
\end{figure}

If the intuitions that drive \ac{ADR} are correct, we should see increased benefit of a learned sampling strategy with larger $N_{rand}$ due to the increasing sparsity of \textit{informative} environments when sampling uniformly. 
We first explore \ac{ADR}'s performance on \texttt{Pusher3Dof-v0}, an environment where $N_{rand}=2$. 
Both randomization dimensions (puck damping, puck friction loss) affect whether or not the puck retains momentum and continues to slide after making contact with the agent's end effector. 
Lowering the values of these parameters \textit{simultaneously} creates an intuitively-harder environment, where the puck continues to slide after being hit. 
In the reference environment, the puck retains no momentum and must be continuously pushed in order to move. 
We qualitatively visualize the effect of these parameters on puck sliding in Figure \ref{pushersampling}.

From Figure \ref{pusheradvantage}, we see \ac{ADR}'s improved robustness to \textit{extrapolation} - or when the target domain lies \textit{outside the training region}. 
We train two agents, one using \ac{ADR} and one using \ac{UDR}, and show them only the training regions encapsulated by the dark, outlined box in the top-right of Figure \ref{pushersampling}.
Qualitatively, only 25\% of the environments have dynamics which cause the puck to slide, which are the hardest environments to solve in the training region. 
We see that from the sampling histogram overlaid on Figure \ref{pushersampling} that \ac{ADR} prioritizes the single, harder purple region more than the light blue regions, allowing for better generalization to the unseen test domains, as shown in Figure \ref{pusheradvantage}. 
\ac{ADR} outperforms \ac{UDR} in all but one test region and produces policies with less variance than their \ac{UDR} counterparts. 

% \begin{figure}[ht]
%   \subfigure[]{% 
%     \includegraphics[width=.475\columnwidth]{plots2/learning_curves_pusher_ref_mean.png} \label{pusherreference} 
%   } 
%   ~ 
%   \subfigure[]{% 
%     \includegraphics[width=.475\columnwidth]{plots2/learning_curves_pusher_hard_mean.png} \label{pusherhard} 
%   } 
%   \caption{Learning curves over time in \textbf{ErgoPusher}. Lower  is  better. \textbf{(a)} Performance  on  the  default  environment settings; \textbf{(b)} Performance on particularly difficult settings - our approach outperforms both the policy trained with the UDR approach both in terms of performance and variance.} 
%   \label{pusherlc}
% \end{figure}

\subsection{Randomization in \textit{Uninterpretable} Dimensions}
\label{sec:ergosim}
We further show the significance of \ac{ADR} over \ac{UDR} on \texttt{ErgoReacher-v0}, where $N_{rand}=8$. 
It is now impossible to infer intuitively which environments are \textit{hard} due to the complex interactions between the eight randomization parameters (gains and maximum torques for each joint). 
For demonstration purposes, we test extrapolation by creating a held-out target environment with extremely low values for torque and gain, which causes certain states in the environment to lead to \textit{catastrophic} failure - gravity pulls the robot end effector down, and the robot is not strong enough to pull itself back up. 
We show an example of an agent getting trapped in a catastrophic failure state in Figure \ref{cfs}, Appendix \ref{app:cfs}.

To generalize effectively, the sampling policy should prioritize environments with lower torque and gain values in order for the agent to operate in such states precisely. 
However, since the hard evaluation environment is not seen during training, \ac{ADR} must learn to prioritize the hardest environments that it can see, while still learning behaviors that can operate well across the entire training region.

% In Figure \ref{ergohard}, we see that when evaluated on the unseen, difficult (low-torque) environment, the policy learned using \ac{ADR} has much lower variance than one learned using \ac{UDR} and can solve the environment much more effectively.

\begin{wrapfigure}{R}{0.4\columnwidth}
  \vspace{-30pt}
  \hspace*{-.75\columnsep}
  \subfigure[]{% 
    \includegraphics[width=.4\columnwidth]{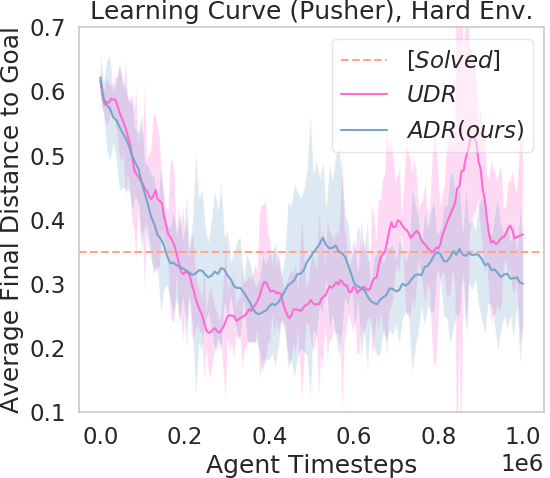} \label{pusherhard} 
  } 
  ~
  \hspace*{-.75\columnsep}
  \subfigure[]{% 
    \includegraphics[width=.45\columnwidth]{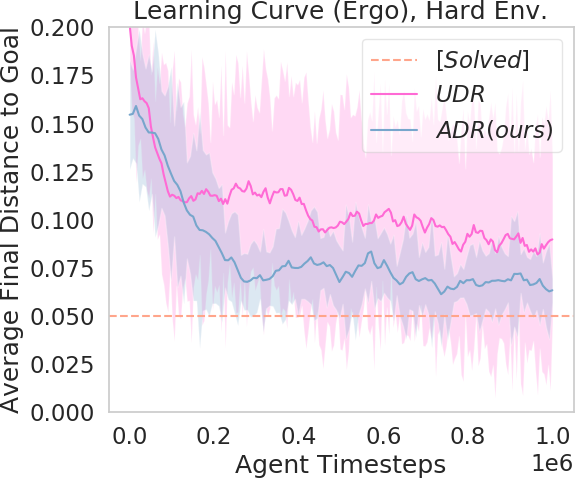} \label{ergohard} 
  } 
  \caption{Learning curves over time in \textbf{(a)} \textbf{Pusher3Dof-v0} and \textbf{(b)} \textbf{ErgoReacher} on held-out, difficult environment settings. Our approach outperforms both the policy trained with the UDR approach both in terms of performance and variance.} 
  \vspace{-40pt}
  \label{ergo}
\end{wrapfigure}

From Figure \ref{pusherhard} (learning curves for \texttt{Pusher3Dof-v0} on the unseen, hard environment - the pink square in Figure \ref{pushergrid}) and \ref{ergohard} (learning curves for \texttt{ErgoReacher-v0} on unseen, hard environment), we observe the detrimental effects of uniform sampling. In \texttt{Pusher3Dof-v0}, we see that \ac{UDR} \textit{unlearns} the good behaviors it acquired in the beginning of training.
When training neural networks in both supervised and reinforcement learning settings, this phenomenon has been dubbed as \textit{catastrophic forgetting}~\cite{Kirkpatrick}. 
\ac{ADR} seems to exhibit this slightly (leading to "hills" in the curve), but due to the adaptive nature the algorithm, it is able to adjust quickly and retain better performance across all environments.

\ac{UDR}'s high variance on \texttt{ErgoReacher-v0} highlights another issue: by continuously training on a random mix of hard and easy \ac{MDP} instances, both beneficial and detrimental agent behaviors can be learned and unlearned throughout training. This mixing can lead to high-variance, inconsistent, and unpredictable behavior upon transfer. 
By focusing on those harder environments and allowing the definition of \textit{hard} to adapt over time, \ac{ADR} shows more consistent performance and better overall generalization than \ac{UDR} in all environments tested.

\subsection{Sim2Real Transfer Experiments}

% In this section, we present results of simulation-trained policies transferred zero-shot onto the real Poppy Ergo Jr. robot.

In \textit{sim2real} (simulation to reality) transfer, many policies fail due to unmodeled dynamics within the simulators, as policies may have overfit to or exploited simulation-specific details of their training environments.
While the deficiencies and high variance of \ac{UDR} are clear even in simulated environments, one of the most impressive results of domain randomization was zero-shot transfer out of simulation onto robots.  
However, we find that the same issues of unpredictable performance apply to \ac{UDR}-trained policies in the real world as well.

We take each method's (\ac{ADR} and \ac{UDR}) five independent simulation-trained policies on both \texttt{ErgoReacher-v0} and \texttt{ErgoPusher-v0} and transfer them without fine tuning onto the real robot. 
We rollout only the final policy on the robot, and show performance in Figure \ref{ergoreal}. 
To evaluate generalization, we alter the environment manually: on \texttt{ErgoReacher-v0}, we change the values of the torques (higher torque means the arm moves at higher speed and accelerates faster); on \texttt{ErgoPusher-v0}, we change the friction of the sliding puck (slippery or rough). For each environment, we evaluate each of the policies with 25 random goals (125 independent evaluations per method per environment setting).

\begin{wrapfigure}{L}{0.35\columnwidth}
%   \vspace{-25pt}
  \hspace*{-.5\columnsep}
  \subfigure[]{% 
    \includegraphics[width=.35\columnwidth]{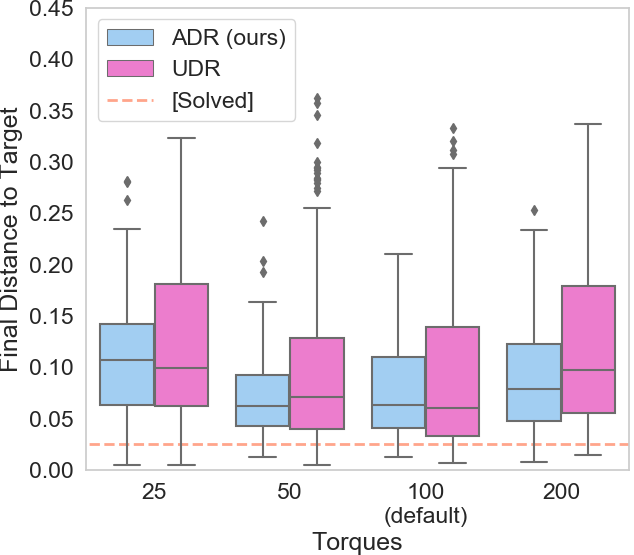} \label{pusherreal} 
  } 
  ~
%   \hspace*{-.2\columnsep} % w/o this, they are horizontally centered
  \subfigure[]{% 
    \includegraphics[width=.30\columnwidth]{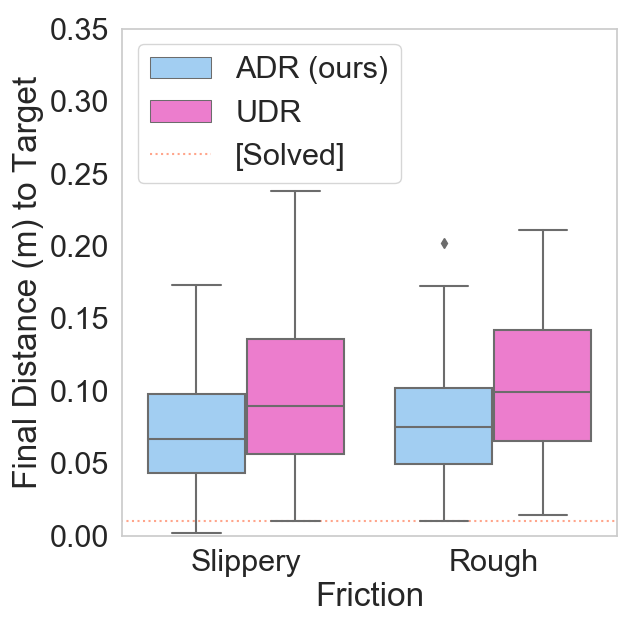} \label{ergoreal} 
  } 
  \caption{Zero-shot transfer onto real robots \textbf{(a)} \textbf{ErgoReacher} and \textbf{(b)} \textbf{ErgoPusher}. In both environments, we assess generalization by manually changing torque strength and puck friction respectively.} 
  \vspace{-35pt}
  \label{ergoreal}
\end{wrapfigure}

Even in zero-shot transfer tasks onto real robots, \ac{ADR} policies obtain overall better or similar performance than \ac{UDR} policies trained in the same conditions.
More importantly, \ac{ADR} policies are more consistent and display lower spread across all environments, which is crucial when safely evaluating reinforcement learning policies on real-world robots.

\section{Related Work}
\subsection{Dynamic and Adversarial Simulators}
Simulators have played a crucial role in transferring learned policies onto real robots, and many different strategies have been proposed. 
Randomizing simulation parameters for better generalization or transfer performance is a well-established idea in evolutionary robotics \cite{zagal2004back,bongard2004once}, but recently has emerged as an effective way to perform zero-shot transfer of deep reinforcement learning policies in difficult tasks \cite{openai2018learning, tobin2017domain, dynrand, SadeghiL16}.

Learnable simulations are also an effective way to adapt a simulation to a particular target environment. \citet{chebotar2018closing} and \citet{ruiz2018learning2sim} use \ac{RL} for effective transfer by learning parameters of a simulation that accurately describes the target domain, but require the target domain for reward calculation, which can lead to overfitting.
In contrast, our approach requires no target domain, but rather only a reference domain (the default simulation parameters) and a general range for each parameter. 
\ac{ADR} encourages diversity, and as a result gives the agent a wider variety of experience. In addition, unlike \citet{chebotar2018closing}, our method does not requires carefully-tuned (co-)variances or task-specific cost functions.
Concurrently, \citet{khirodkar2018vadra} also showed the advantages of learning adversarial simulations and disadvantages of purely uniform randomization distributions in object detection tasks. 

To improve policy robustness, \ac{RARL}  \citet{rarl} jointly trains both an agent and an adversary who applies environment forces that disrupt the agent's task progress. 
\ac{ADR} removes the zero-sum game dynamics, which have been known to decrease training stability \cite{mescheder2018training}. 
More importantly, our method's final outputs - the SVPG-based sampling strategy and discriminator - are reusable and can be used to train new agents as shown in Appendix \ref{newagentsscratch}, whereas a trained \ac{RARL} adversary would overpower any new agent and impede learning progress. 

\subsection{Active Learning and Informative Samples}
Active learning methods in supervised learning try to construct a \textit{representative}, sometimes time-variant, dataset from a large pool of unlabelled data by proposing elements to be labeled. The chosen samples are labelled by an oracle and sent back to the model for use. Similarly, \ac{ADR} searches for what environments may be most useful to the agent at any given time. Active learners, like \ac{BO} methods discussed in Section \ref{sec:method}, often require an acquisition function (derived from a notion of model uncertainty) to chose the next sample. Since \ac{ADR} handles this decision through the  explore-exploit framework of \ac{RL} and the $\alpha$ in \ac{SVPG}, \ac{ADR} sidesteps the well-known scalability issues of both active learning and \ac{BO} \cite{tong2001}. 

Recently, \citet{toneva2018empirical} showed that certain examples in popular computer vision datasets are harder to learn, and that some examples are forgotten much quicker than others. We explore the same phenomenon in the space of \ac{MDP}s defined by our randomization ranges, and try to find the "examples" that cause our agent the most trouble. Unlike in active learning or \citet{toneva2018empirical}, we have no oracle or supervisory loss signal in \ac{RL}, and instead attempt to learn a proxy signal for \ac{ADR} via a discriminator.
\subsection{Generalization in Reinforcement Learning}
Generalization in RL has long been one of the holy grails of the field, and recent work like \citet{packer2018generalization}, \citet{Cobbe2018QuantifyingGI}, and \citet{farebrother2018generalization} highlight the tendency of deep RL policies to overfit to details of the training environment. Our experiments exhibit the same phenomena, but our method improves upon the state of the art by explicitly searching for and varying the environment aspects that our agent policy may have overfit to. 
We find that our agents, when trained more frequently on these problematic samples, show better generalization over all environments tested.
\section{Conclusion}

In this work, we highlight failure cases of traditional domain randomization, and propose active domain randomization (ADR), a general method capable of finding the most informative parts of the randomization parameter space for a reinforcement learning agent to train on.
ADR does this by posing the search as a reinforcement learning problem, and optimizes for the most informative environments using a learned reward and multiple policies.
We show on a wide variety of simulated environments that this method efficiently trains agents with better generalization than traditional domain randomization, extends well to high dimensional parameter spaces, and produces more robust policies when transferring to the real world. 

% \section*{Acknowledgements}
% The authors gratefully acknowledge the Natural Sciences and Engineering Research Council of Canada (NSERC), the Fonds de Recherche Nature et Technologies Quebec (FQRNT) and the Open Philanthropy Project for supporting this work. In addition, the authors would like to thank Kyle Kastner and members of the REAL Lab for their helpful comments.

\bibliography{example}
\bibliographystyle{corlabbrvnat}

\clearpage
\begin{appendices}

\section{Architecture Walkthrough}
\label{sec:walkthrough}

In this section, we walk through the diagram shown in Figure \ref{fig:overview}. All line references refer to Algorithm \ref{alg:adr}.

\subsubsection{SVPG Sampler}
To encourage sufficient exploration in high dimensional randomization spaces, we parameterize $\mu_\phi$ with \ac{SVPG}. Since each particle proposes its own environment settings $\xi_i$ (lines 4-6, Figure \ref{fig:overview}h), all of which are passed to the agent for training, the agent policy benefits from the same environment variety seen in \ac{UDR}. However, unlike \ac{UDR}, $\mu_\phi$ can use the learned reward to focus on problematic MDP instances while still being efficiently parallelizable.

\subsubsection{Simulator}
After receiving each particle's proposed parameter settings $\xi_i$, we generate randomized environments $E_i = S(\xi_i)$ (line 9, Figure  \ref{fig:overview}b).

\subsubsection{Generating Trajectories}

We proceed to train the agent policy $\pi$ on the randomized instances $E_i$, just as in \ac{UDR}.
We roll out $\pi$ on each randomized instance $E_i$ and store each trajectory $\tau_i$.
For every randomized trajectory generated, we use the \textit{same} policy to collect and store a reference trajectory $\tau_{ref}$ by rolling out $\pi$ in the default environment $E_{ref}$ (lines 10-12, Figure  \ref{fig:overview}a, c).
We store all trajectories (lines 11-12) as we will use them to score each parameter setting $\xi_i$ and update the discriminator.

The agent policy is a \textit{black box}: although in our experiments we train $\pi$ with Deep Deterministic Policy Gradients \cite{lillicrap2015continuous}, the policy can be trained with any other on or off-policy algorithm by introducing only minor changes to Algorithm \ref{alg:adr} (lines 13-17, Figure  \ref{fig:overview}d). 

\subsubsection{Scoring Environments}
We now generate a score for each environment (lines 20-22) using each stored randomized trajectory $\tau_i$ by passing them through the discriminator $D_{\psi}$, which predicts the type of environment (reference or randomized) each trajectory was generated from.
We use this score as a reward to update each \ac{SVPG} particle using Equation \ref{eq:svpg} (lines 24-26, Figure  \ref{fig:overview}f).

After scoring each $\xi_i$ according to Equation \ref{eq:discrimrew}, we use the randomized and reference trajectories to train the discriminator (lines 28-30, Figure  \ref{fig:overview}e).

\section{Learning Curves for Reference Environments}

For space concerns, we show only the \textit{hard} generalization curves for all environments in the main document. For completeness, we include learning curves on the reference environment here.

\begin{figure}[h]
  \centering
  \subfigure[]{% 
    \includegraphics[width=.3\columnwidth]{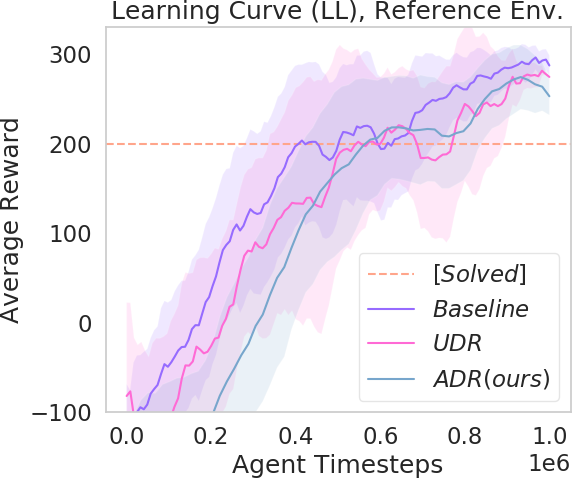}\label{lcreferencell} 
  } 
  ~
  \subfigure[]{% 
    \includegraphics[width=.3\columnwidth]{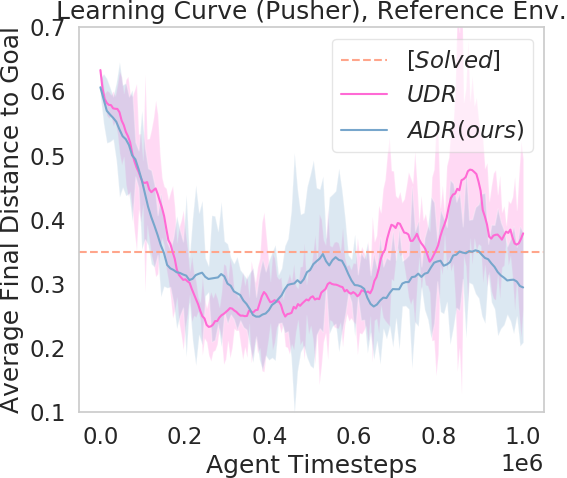} \label{lcreferencepusher} 
  } 
  ~ 
  \subfigure[]{% 
    \includegraphics[width=.3\columnwidth]{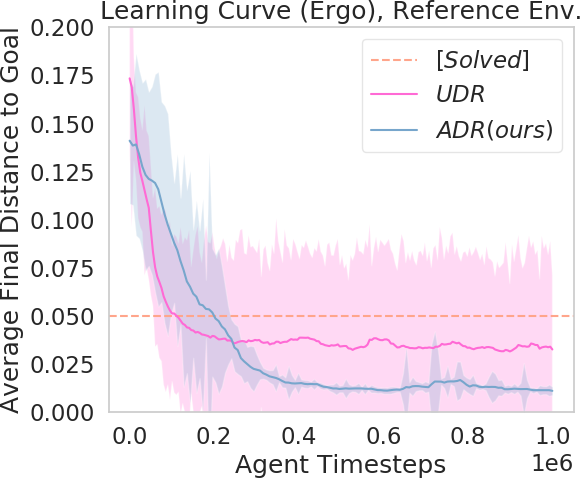}\label{lcreferenceergo} 
  } 
  \caption{Learning curves over time reference environments.
  \textbf{(a)} LunarLander \textbf{(b)} Pusher-3Dof \textbf{(c)} ErgoReacher.} 
%   \label{lunarours}
% \end{figure}
% \begin{figure}[t]
%   \caption{ \textbf{(b)} Change in dynamics sampling during training. As training proceeds, ADR begins preferentially sampling the more challenging environmental instances.} 
\end{figure}

\section{Interpretability Benefits of ADR}
\begin{figure}
\centering
\includegraphics[width=0.5\columnwidth]{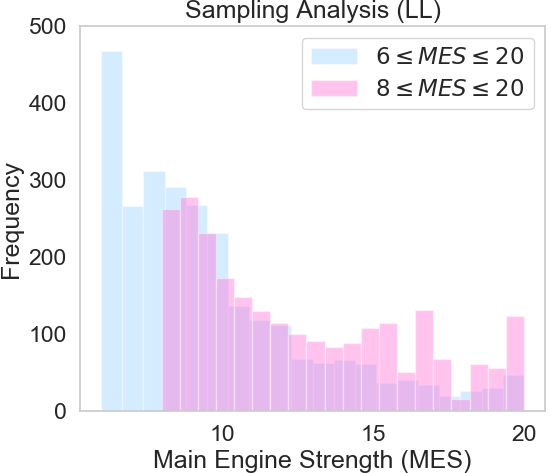}
  \caption{Sampling frequency across engine strengths when varying the randomization ranges. The updated, red distribution shows a much milder unevenness in the distribution, while still learning to focus on the harder instances. 
%   This can be used for debugging the randomization ranges before transferring a learned policy onto a physical system.} 
}
  \label{debugsim}
\end{figure}

One of the secondary benefits of ADR is its insight into incompatibilities between the task and randomization ranges. 
We demonstrate the simple effects of this phenomenon in a one-dimensional \texttt{LunarLander-v2}, where we only randomize the main engine strength. 
Our initial experiments varied this parameter between 6 and 20, which lead to ADR learning degenerate agent policies by learning to propose the lopsided blue distribution in Figure \ref{debugsim}. 
Upon inspection of the simulation, we see that when the parameter has a value of less than approximately 8, the task becomes almost impossible to solve due to the other environment factors (in this case the lander always hits the ground too fast, which it is penalized for). 

After adjusting the parameter ranges to more sensible values, we see a better sampled distribution in pink, which still gives more preference to the hard environments in the lower engine strength range. 
Most importantly, ADR allows for analysis that is both \textit{focused} - we know exactly what part of the simulation is causing trouble - and \textit{pre-transfer}, i.e. done before a more expensive experiment such as real robot transfer has taken place. 
With \ac{UDR}, the agents would be equally trained on these degenerate environments, leading to policies with potentially undefined behavior (or, as seen in Section \ref{sec:ergosim}, unlearn good behaviors) in these truly out-of-distribution simulations.

\section{Bootstrapping Training of New Agents}\label{newagentsscratch}
\begin{figure}[h]
  \subfigure[]{% 
    \includegraphics[width=.475\columnwidth]{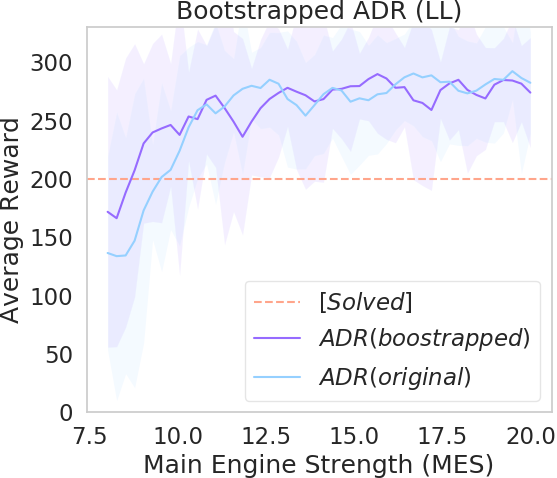} \label{bootstrapgen} 
  } 
  ~ 
  \subfigure[]{% 
    \includegraphics[width=.475\columnwidth]{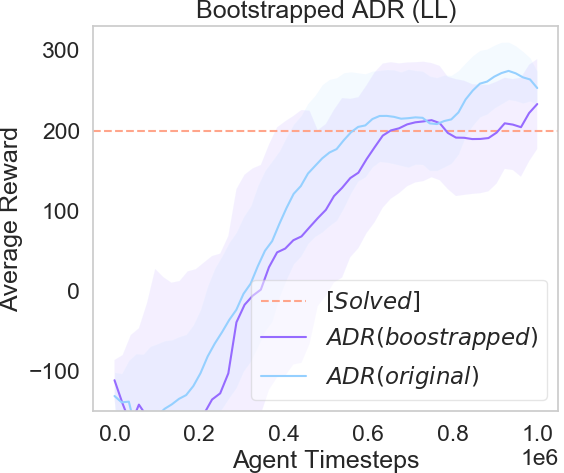} \label{bootstraplc} 
  } 
  \caption{Generalization and default environment learning progression on \texttt{LunarLander-v2} when using ADR to bootstrap a new policy. Higher is better.} 
  \label{bootstrap}
\end{figure}

Unlike \ac{DR}, \ac{ADR}'s learned sampling strategy and discriminator can be reused to train new agents from scratch. 
To test the transferability of the sampling strategy, we first train an instance of \ac{ADR} on \texttt{LunarLander-v2}, and then extract the \ac{SVPG} particles and discriminator. 
We then replace the agent policy with an random network initialization, and once again train according the the details in Section \ref{implementation}. 
From Figure \ref{bootstrapgen}, it can be seen that the bootstrapped agent generalization is even better than the one learned with \ac{ADR} from scratch. 
However, its training speed on the default environment ($\xi_{MES}=13$) is relatively lower.

\section{Uniform Domain Randomization}
\label{app:udr}
Here we review the algorithm for Uniform Domain Randomization (UDR), first proposed in \cite{tobin2017domain}, shown in Algorithm \ref{alg:dr}.

\begin{algorithm}[h]
   \caption{Uniform Sampling Domain Randomization}
   \label{alg:dr}
\begin{algorithmic}[1]
      \STATE {\textbf{Input}: $\Xi$: Randomization space, $S$: Simulator} 
   \STATE \textbf{Initialize} $\pi_\theta$: agent policy
   \FOR{each episode}
   \STATE // \textit{Uniformly sample parameters}
   \FOR{$i=1$ {\bfseries to} $N_{rand}$}
   \STATE $\xi^{(i)} \sim U\big[ \xi^{(i)}_{low}, \xi^{(i)}_{high} \big]$
   \ENDFOR
   \STATE { \textit{// Generate, rollout in randomized env.}} \label{l:start-rollout2}
   \STATE $E_i \leftarrow S(\xi_i)$ 
   \STATE \textbf{rollout} $\tau_i \sim \pi_\theta(\cdot; E_{i})$
   \STATE{$\mathcal{T}_{rand} \gets \mathcal{T}_{rand} \cup \tau_i$}
   \FOR{ \textbf{each} gradient step}
   \STATE  \textit{// Agent policy update}
   \STATE \textbf{with} $\mathcal{T}_{rand}$ \textbf{update}:
   \STATE $\quad \theta \leftarrow \theta + \nu\nabla_{\theta}J(\pi_\theta)$
   \ENDFOR
   \ENDFOR
   
\end{algorithmic}
\end{algorithm}

\section{Environment Details}
\label{app:envtable}
Please see Table \ref{table:envtable}.

\subsection{Catastrophic Failure States in ErgoReacher}
\label{app:cfs}
\begin{figure}[h]
\begin{center}
\centerline{\includegraphics[width=\columnwidth]{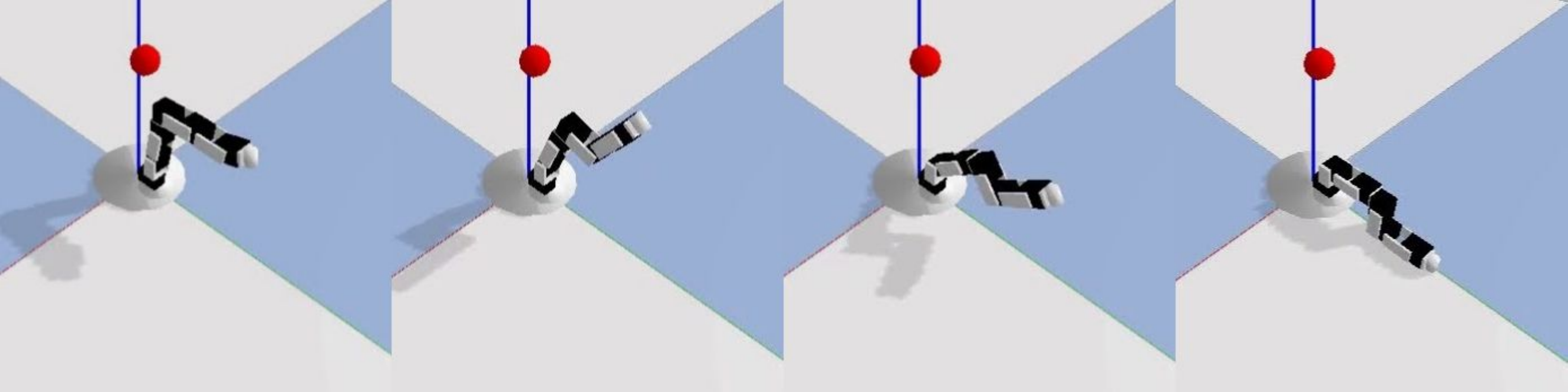}}
\caption{An example progression (left to right) of an agent moving to a catastrophic failure state (Panel 4) in the hard \texttt{ErgoReacher-v0} environment.}
\label{cfs}
\end{center}
\end{figure}

In Figure \ref{cfs}, we show an example progression to a \textit{catastrophic failure state} in the held-out, simulated target environment of \texttt{ErgoReacher-v0}, with extremely low torque and gain values.

\begin{table*}[t]
\renewcommand{\arraystretch}{1.2}
\begin{tabular}{@{}lllll@{}}
\toprule
Environment & $N_{rand}$ & Types of Randomizations & Train Ranges & Test Ranges \\ \midrule
\texttt{LunarLander-v2} & 1  & Main Engine Strength & $[8, 20]$ & $[8, 11]$ \\
\texttt{Pusher-3DOF-v0} & 2  & Puck Friction Loss \& Puck Joint Damping & $[0.67, 1.0] \times$ default & $[0.5, 0.67] \times$ default \\
\texttt{ErgoPusher-v0} & 2  & Puck Friction Loss \& Puck Joint Damping & $[0.67, 1.0] \times$ default & $[0.5, 0.67] \times$ default \\
\multirow{2}{*}{\texttt{ErgoReacher-v0}} & \multirow{2}{*}{8}  &     Joint Damping   & $[0.3, 2.0] \times$ default & $0.2 \times$ default \\ 
 & & Joint Max Torque & $[1.0, 4.0] \times$ default & default \\ \bottomrule
\end{tabular}
\caption{We summarize the environments used, as well as characteristics about the randomizations performed in each environment.}
\label{table:envtable}
\end{table*}

\section{Untruncated Plots for Lunar Lander}
\begin{figure}[h]
    \centering
    \includegraphics[width=0.7\columnwidth]{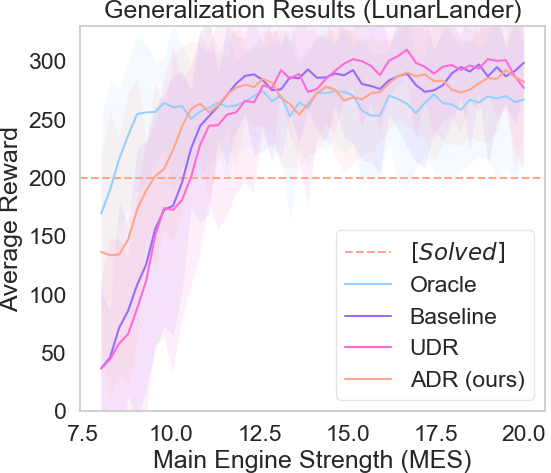}
    \caption{Generalization on \texttt{LunarLander-v2} for an expert interval selection, \ac{ADR}, and \ac{UDR}. Higher is better.}
    \label{fig:adrgeneralizationfull}
\end{figure}

All policies on Lunar Lander described in our paper receive a \textit{Solved} score when the engine strengths are above 12, which is why truncated plots are shown in the main document. For clarity, we show the full, untruncated plot in Figure \ref{fig:adrgeneralizationfull}.

\section{Network Architectures and Experimental Hyperparameters}
\label{app:expdetails}
All experiments can be reproduced using our Github repository\footnotemark.
\footnotetext{Code link will be updated after review.}
% \footnotetext{\href{https://github.com/montrealrobotics/active-domainrand}{https://github.com/montrealrobotics/active-domainrand}}

All of our experiments use the same network architectures and experiment hyperparameters, except for the number of particles $N$. For any experiment with \texttt{LunarLander-v2}, we use $N=10$. For both other environments, we use $N=15$. All other hyperparameters and network architectures remain constant, which we detail below. All networks use the Adam optimizer \cite{kingma2014adam}.

We run Algorithm \ref{alg:adr} until 1 million \textit{agent timesteps} are reached - i.e. the agent policy takes 1M steps in the randomized environments. We also cap each episode off a particular number of timesteps according to the documentation associated with \cite{brockman2016gym}. In particular, \texttt{LunarLander-v2} has an episode time limit of 1000 environment timesteps, whereas both \texttt{Pusher-3DOF-v0} and \texttt{ErgoReacher-v0} use an episode time limit of 100 timesteps.

For our agent policy, we use an implementation of DDPG (particularly, \texttt{OurDDPG.py}) from the Github repository associated with \cite{Fujimoto2018AddressingFA}. The actor and critic both have two hidden layers of 400 and 300 neurons respectively, and use \texttt{ReLU} activations. Our discriminator-based rewarder is a two-layer neural network, both layers having 128 neurons.  The hidden layers use \texttt{tanh} activation, and the network outputs a \texttt{sigmoid} for prediction. 

The agent particles in SVPG are parameterized by a two-layer actor-critic architecture, both layers in both networks having 100 neurons. We use \ac{A2C} to calculate unbiased and low variance gradient estimates. All of the hidden layers use \texttt{tanh} activation and are orthogonally initialized, with a learning rate of $0.0003$ and discount factor $\gamma = 0.99$. They operate on a $\mathbb{R}^{N_{rand}}$ continuous space, with each axis bounded between $[0, 1]$. We allow for set the max step length to be 0.05, and every 50 timesteps, we reset each particle and randomly initialize its state using a ${N_{rand}}$-dimensional uniform distribution. We use a temperature $\alpha = 10$ with an RBF-Kernel as was done in \cite{svpg}. In our work we use an \ac{RBF} kernel with median baseline as described in \citet{svpg} and an \ac{A2C} policy gradient estimator \cite{mnih2016asynchronous}, although both the kernel and estimator could be substituted with alternative methods \cite{gangwani2018diverse}. To ensure diversity of environments throughout training, we always roll out the SVPG particles using a non-deterministic sample. 

For DDPG, we use a learning rate $\nu = 0.001$, target update coefficient of $0.005$, discount factor $\gamma = 0.99$, and batch size of 1000. We let the policy run for 1000 steps before any updates, and clip the max action of the actor between $[-1, 1]$ as prescribed by each environment.

Our discriminator-based reward generator is a network with two, 128-neuron layers with a learning rate of $.0002$ and a binary cross entropy loss (i.e. is this a \textit{randomized} or \textit{reference} trajectory). To calculate the reward for a trajectory for any environment, we split each trajectory into its $(s_t, a_t, s_{t+1})$ constituents, pass each tuple through the discriminator, and average the outputs, which is then set as the reward for the trajectory. Our batch size is set to be 128, and most importantly, as done in \cite{eysenbach2018diversity}, we calculate the reward for examples before using those same examples to train the discriminator.

\end{appendices}

\end{document}